\documentclass[a4paper,fleqn]{cas-dc}

\usepackage[numbers]{natbib}
\usepackage{caption}
\usepackage{indentfirst}
\usepackage{graphicx}
\usepackage{float}
\usepackage{flushend}
\usepackage{stfloats}
\usepackage{booktabs} 
\usepackage{diagbox} 
\usepackage{tabu}
\usepackage{amsmath}
\usepackage{array}
\usepackage{color}
\newcommand{\etal}{{\emph{et al.}}}
\usepackage{gensymb}
\usepackage{multirow}
\usepackage{threeparttable}
\usepackage{bbding}
\usepackage{pifont}
\usepackage{color,xcolor}

\usepackage{soul}
\soulregister\cite7
\soulregister\ref7

\ExplSyntaxOn
\keys_set:nn { stm / mktitle } { nologo }
\ExplSyntaxOff

\begin{document}
\let\WriteBookmarks\relax
\def\floatpagepagefraction{1}
\def\textpagefraction{.001}
\shorttitle{}
\shortauthors{Yilan Zhang et~al.}

\title [mode = title]{TFormer: A throughout fusion transformer for multi-modal skin lesion diagnosis}                      



\author[1]{Yilan Zhang}
\author[1]{Fengying Xie}
\cormark[1] 
\cortext[cor1]{Corresponding author}
\ead{xfy_73@buaa.edu.cn}
\author[1]{Jianqi Chen}
\address[1]{Image Processing Center, School of Astronautics, Beihang University, Beijing 100191, China}

\begin{abstract}
Multi-modal skin lesion diagnosis (MSLD) has achieved remarkable success by modern computer-aided diagnosis (CAD) technology based on deep convolutions. However, the information aggregation across modalities in MSLD remains challenging due to severity unaligned spatial resolution (e.g., dermoscopic image and clinical image) and heterogeneous data (e.g., dermoscopic image and patients' meta-data). Limited by the intrinsic local attention, most recent MSLD pipelines using pure convolutions struggle to capture representative features in shallow layers, thus the fusion across different modalities is usually done at the end of the pipelines, even at the last layer, leading to an insufficient information aggregation. To tackle the issue, we introduce a pure transformer-based method, which we refer to as "Throughout Fusion Transformer (TFormer)", for sufficient information integration in MSLD. Different from the existing approaches with convolutions, the proposed network leverages transformer as feature extraction backbone, bringing more representative shallow features. We then carefully design a stack of dual-branch hierarchical multi-modal transformer (HMT) blocks to fuse information across different image modalities in a stage-by-stage way. With the aggregated information of image modalities, a multi-modal transformer post-fusion (MTP) block is designed to integrate features across image and non-image data. Such a strategy that information of the image modalities is firstly fused then the heterogeneous ones enables us to better divide and conquer the two major challenges while ensuring inter-modality dynamics are effectively modeled. Experiments conducted on the public Derm7pt dataset validate the superiority of the proposed method. Our TFormer achieves an average accuracy of 77.99\%  and diagnostic accuracy of 80.03\% , which outperforms other state-of-the-art methods. Ablation experiments also suggest the effectiveness of our designs. The codes can be publicly available from \url{https://github.com/zylbuaa/TFormer.git}.
\end{abstract}

\begin{graphicalabstract}
\includegraphics[scale=0.5]{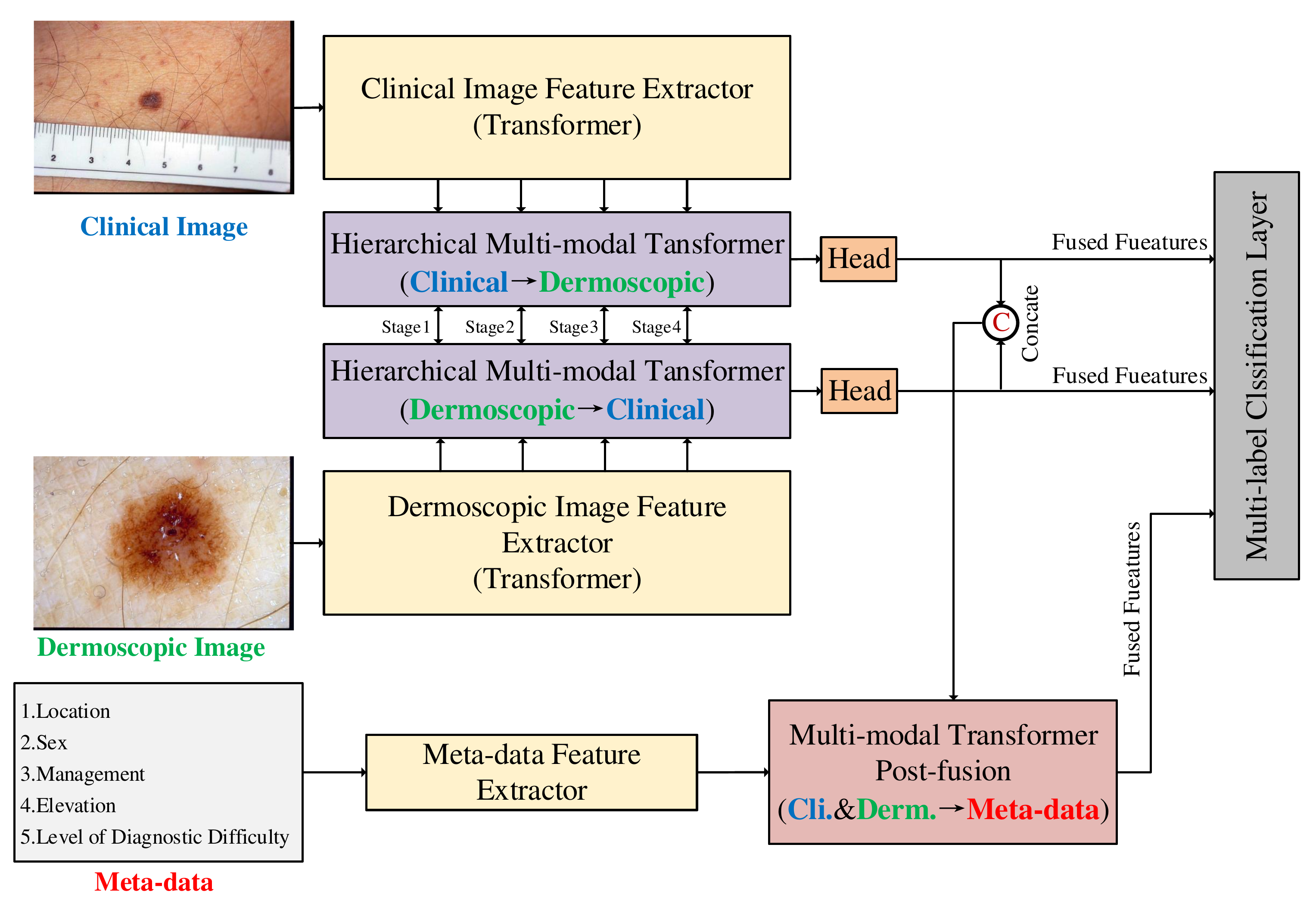}
\end{graphicalabstract}

\begin{highlights}
\item We propose a pure transformer, TFormer, for multi-modal skin lesion diagnosis. 
\item We design a dual-branch HMT block for image modalities fusion. 
\item A ``divide and conquer'' strategy is adopted to tackle fusions between modalities.
\item The proposed TFormer achieves state-of-the-art performance on a benchmark dataset.
\end{highlights}


\maketitle

\section{Introduction}
Skin cancer is one of the most common types of cancer by far \cite{balch2009final}. Some types of skin cancer, such as melanoma, can easily spread out to other body parts if not caught and treated in time \cite{skincaner}, thus have a great danger. To enable an accurate and confident skin lesion diagnosis (SLD), the recent approaches resort to computer-aided diagnosis (CAD) technology by leveraging powerful discriminative capability of deep learning. Imitating the routine clinical diagnosis \cite{tang2022fusionm4net} where dermatologists make final assessment based on different data modalities, e.g., patients' diagnostic images and medical records, Kawahara \etal \cite{kawahara2018seven} proposed Derm7pt, a multi-modal dataset containing three modalities (dermoscopic images, clinical images, and patients' meta-data), which facilitates the research on multi-modal skin lesion diagnosis (MSLD). Dermoscopic images reveal local structures beneath the skin's surface, while clinical images focus on global information, and meta-data are often used as a supplementary reference. Compared with using a single modality, MSLD enables deep networks better integrate complementary information from diverse modalities and provide more robust predictions for diagnosis, thus making it a hot research topic.

\begin{figure*}
	\centering
	\includegraphics[scale=.8]{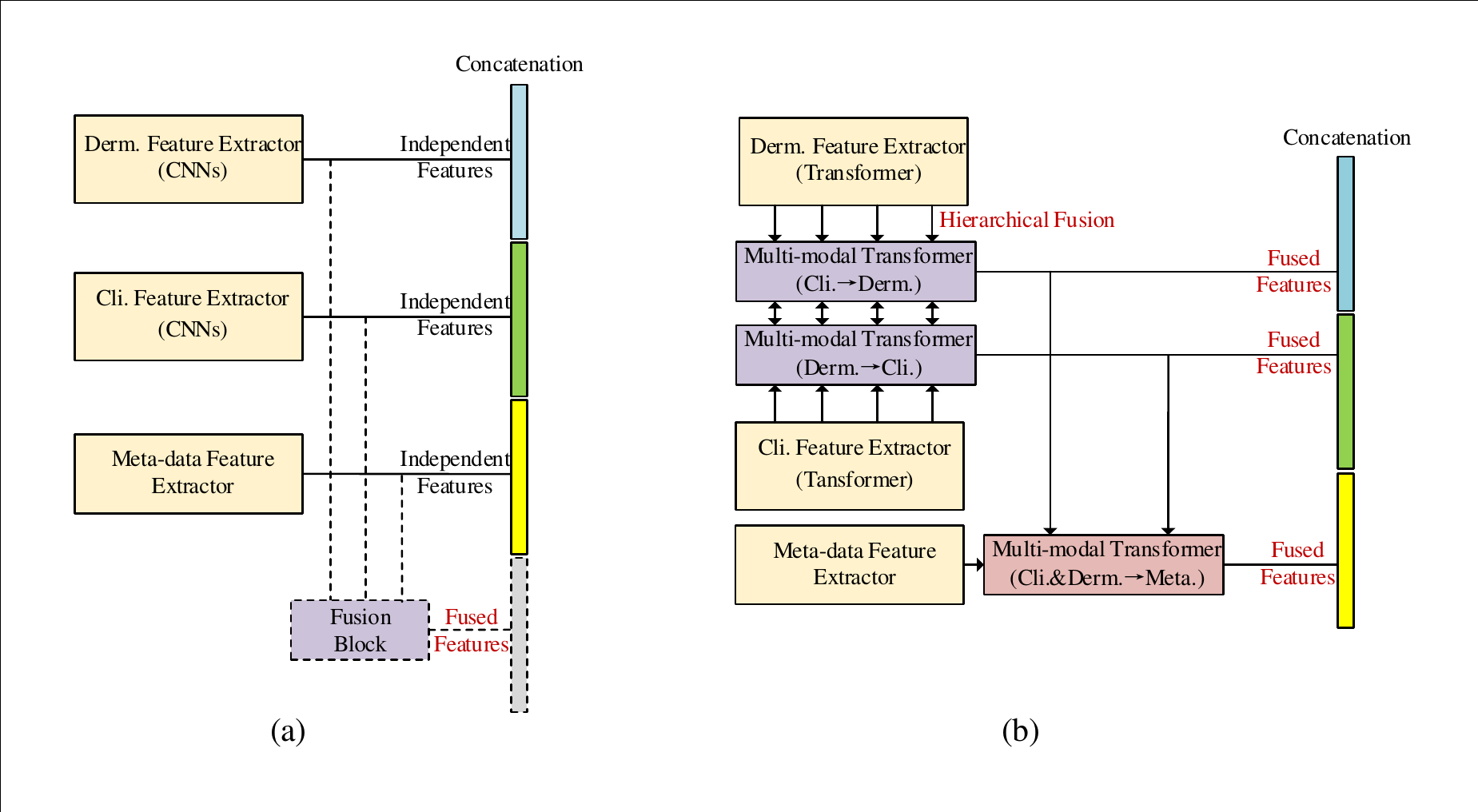}
	\caption{Different architectures for multi-modal skin lesion classification. (a)The majority of CNN-based methods. (b) Our transformer-based method.}
	\label{architecture}
\end{figure*}

Existing MSLD methods are primarily based on deep convolutions neural networks (CNN) \cite{ge2017skin,kawahara2018seven,yap2018multimodal, nunnari2020study,nedelcu2020multi,wang2021interpretability,lu2021multimodal,tang2022fusionm4net}. Most of them implement a multi-modal framework similar to that shown in Figure \ref{architecture} (a). Features of image modalities are extracted by CNNs such as VGGs \cite{simonyan2014very}, Inception-V3 \cite{szegedy2016rethinking}, ResNets \cite{he2016deep}, while features of meta-data are obtained by one-hot encoding or multilayer perceptron (MLP). These independently extracted features are concatenated directly or fused via a fusion block before being fed into the classification layer. The fusion blocks are usually several fully connected layers or machine learning methods such as random forest \cite{nunnari2020study}, which are typically located in the last few layers of CNNs. Although these methods have achieved notable success in MSLD, there remains two defects.

First, the information fusion across image modalities is insufficient. Due to the different imaging mechanisms and standards, there are various scales of the same skin lesion in diverse modalities. Take images in Derm7pt as an example, as the skin lesion regions fulfill the whole dermoscopic images while partial clinical images, the two modalities differ much from the perspective of spatial resolution (we refer to it as "unaligned spatial resolution"). Owing to the intrinsic locality of convolution operations, the shallow layers of CNNs struggle to extract contextual information from the image modalities. Therefore, when the images processed by shallow convolution layers, the extracted features are still severe unaligned, causing negligible benefits of the information fusion in this part. Besides, many existing methods \cite{ge2017skin, nunnari2020study, tang2022fusionm4net} just apply the fusion at the end layers of networks and even the last decision layer, which can lead to insufficient information intercorrelation modeling, as the spatial information is highly condensed and some useful information is lost. 

Second, there lacks attention on the severe gap between different data types across modalities (we refer to it as "heterogeneous data"). Current MSLD methods either directly concatenate the image and meta-data features at the classification layer \cite{kawahara2018seven}, or feed the concatenated features into several MLPs for a simple fusion \cite{yap2018multimodal}. With features of each modality extracted independently as illustrated in Figure \ref{architecture} (a), such strategies actually regard image modality and meta-data modality as equal which may lead to sub-optimal balance of information fusion across modalities with different data types.

Inspired by the newly emerging transformer structure that has achieved competitive performance on a variety of visual tasks due to its superiority in modeling long-range dependencies \cite{han2022survey}, in this paper, we propose a pure transformer-based network, TFormer, to tackle the above two challenges (structure displayed in Figure \ref{TFormer}). Different from the convolution structure, transformer leverages a multi-head self-attention mechanism that covers a much wider receptive field \cite{vaswani2017attention}, bringing the features extracted by the shallow layers more representative. Taking advantage of this good characteristic, we utilize modern transformer structure \cite{liu2021swin} as the backbone to extract features of image modalities, while keeping MLP for meta-data. With much representative shallow features, we design a dual-branch hierarchical feature fusion structure called hierarchical multi-modal transformer (HMT) block for sufficient information fusion across clinical and dermoscopic modalities. The HMT block, with its each branch mainly corresponding to a specific modality, performs a cross-modality attention mechanism between the branches, thus strengthen the bimodal interaction ability of the network. In the TFormer, we apply a stack of HMT blocks in accordance to the stages of the transformer backbone, from shallow layers to deep ones. Such a throughout fusion strategy can sufficiently integrate the information across image modalities, and benefit the final diagnosis accuracy that is verified by the experiments in Section \ref{Experiments}. Considering the great difference between image and meta-data modalities, instead of fusing the three modalities synchronously, here we adopt a "divide and conquer" strategy, that is, first fuse modalities of the same data type, then aggregate the different ones. With the fused features of image modalities after HMT blocks, we design a multi-modal transformer post-fusion (MTP) block to fuse information between image and meta-data modalities. The inter-modality dynamics can be properly modeled using the cross-attention mechanism to ensure effective fusion.

Our TFormer network, unlike most existing methods that only perform multi-modal fusion in the last few layers, performs an end-to-end throughout integrated hierarchical feature fusion framework as shown in Figure \ref{architecture} (b). With a pure transformer-based architecture, the TFormer can effectively achieve sufficient modeling intra diverse modalities, as well as the optimal fusion inter modalities. Extensive experiments conducted on the public Derm7pt dataset have demonstrated the superiority of TFormer compared with other state-of-the-art methods. Further ablation experiments have verified the effectiveness of the designed HMT and MTP blocks.

In summary, the contributions of this research are as follows:

\begin{itemize}
	\item We propose a throughout fusion transformer (TFormer) architecture for MSLD. Unlike most existing methods that rely on deep convolution structure and thus struggle to sufficiently fuse multi-modal information, the TFormer is purely designed based on transformer, performs a throughout sufficient information aggregation, and can well tackle the defects that exist in the current methods.
	
	\item We design a dual-branch hierarchical multi-modal transformer (HMT) block that can effectively fuse the information across dermoscopic and clinical modalities. For modalities of heterogeneous data, we adopt a "divide and conquer" strategy and design a multi-modal transformer post-fusion (MTP) block to better interactively aggregate image and meta-data information.
	
	\item Experimental results on the Derm7pt dataset have demonstrated better diagnosis results of TFormer than other state-of-the-art methods. Further ablation study also demonstrates the effectiveness of each element of TFormer.
\end{itemize}

\section{Related Works}

\subsection{Multi-modal Learning for Skin Disease Diagnosis}
With the development of deep learning technology, a significant amount of research has centered on designing effective CNNs for skin image classification \cite{zhang2019attention,yao2021single,ding2021deep}. Currently, many methods in SLD only consider leveraging single dermoscopic modality \cite{xie2016melanoma,yao2021single}. Although these methods achieve some good results, they are somewhat inconsistent with the routine diagnosis practice of dermatologists, where the final diagnosis is concluded by integrating dermoscopic images, clinical observations, and meta-data from patients. Considering that, Kawahara \etal released a large MSLD dataset Derm7pt \cite{kawahara2018seven} which includes the three modalities mentioned above, which facilitates the research in MSLD and promote the accuracy and credibility of the CAD result.

To leverage the information across different modalities, most of the current MSLD methods adopt a late-fusion architecture at the end of the framework. These methods can be broadly classified into two categories. Methods of the first category \cite{nunnari2020study,wang2021interpretability,tang2022fusionm4net} often introduce machine learning methods, such as SVM and random forest, to concatenate meta-data and features of the image modalities processed by CNNs at the decision level. \cite{wang2021interpretability} proposed to concatenate deep features and hand-crafted features processed with Catboost, and leverage the softmax classifier to make the final classification. \cite{tang2022fusionm4net} proposed a FusionM4Net which focuses on the decision-level fusion between a SVM and a fully connection classifier with an additional step of voting weights search. These methods are often carried out in two or more stages, thus cumbersome. In addition, as the feature fusion mainly performs at the decision layer, the inter-modality complementary information is hard to be modeled.

To effectively exploit the end-to-end superiority of deep learning, the other category methods \cite{ge2017skin,kawahara2018seven,yap2018multimodal,lu2021multimodal,nedelcu2020multi} employ pure CNNs to extract independent features of each modality, and then fuse the features using concatenation. Among them, Ge \etal \cite{ge2017skin} proposed a triplet network for melanoma detection that concatenate the feature maps from two pre-trained VGG16 \cite{simonyan2014very}. Yap \etal \cite{yap2018multimodal} extracted feature vectors from dermoscopic and clinical image modalities using two ResNet50 \cite{he2016deep}, concatenated the embedded meta-data, and fed them into an MLP to classify skin cancer. Kawahara \etal \cite{kawahara2018seven} suggested fine-tuning two Inception-V3 \cite{szegedy2016rethinking} to get the features and splicing with meta-data for the MSLD lesion classification. Although these methods have demonstrated great performance, they still adopt a late-fusion strategy. To tackle the issue, in this paper, we propose an end-to-end throughout fusion architecture which performs information fusion across modalities from shallow layers to deep layers. The work most relevant to ours is probably \cite{bi2020multi}. However, they only consider using image modalities while ignoring the meta-data one, leading to an insufficient utilization of the existing datasets. What's more, they construct their network based on pure CNNs, while ours is purely transformer-based.

\subsection{Transformer-based Modal}
Transformer is a type of neural network that entirely depends on self-attention mechanism and was first applied in the field of natural language processing (NLP) \cite{vaswani2017attention}. With its significant advancement in NLP, researchers began to experiment with applying it to computer vision. Chen \etal \cite{chen2020generative} proved that the transformer can be used for pre-training of image classification tasks. ViT \cite{dosovitskiy2020image}, the pioneer of vision transformer, divides the image into fixed-size patches, and each patch is flattened and passes through several transformer blocks, achieving comparable results to CNNs. However, due to the lack of biased induction, ViT often relies on pre-training on large datasets such as JFT-300M and ImageNet-21k \cite{raghu2021vision}.

Recently, the follow-ups \cite{touvron2021training,han2021transformer,yuan2021tokens,wang2021interpretability,chu2021twins,lee2021vision,liu2021swin,wang2021pyramid} improved ViT for resolving the problems above. Wang \etal \cite{wang2021pyramid} developed a pyramid transformer. They designed spatial-reduction attention (SRA) to process high-resolution feature maps and save computation costs by reducing the spatial dimensions of keys and values. Yuan \etal \cite{yuan2021tokens} integrated adjacent tokens through re-structuring and soft split, effectively shortening token length and enhancing local information. 

Considering that the transformer lacks bias induction, which leads to insufficient detail feature extraction, ViT cannot achieve the compared performance as CNN in the case of the limited skin image datasets\cite{gulzar2022skin}. Thus, some improved transformers were designed. These studies focused on the application of transformers in the field of skin image analysis, which achieved remarkable success in skin image classification \cite{aladhadh2022effective,xin2022improved} and segmentation \cite{wang2021boundary,wu2022fat,he2022fully}. Wang \emph{et al.} \cite{wang2021boundary} proposed a boundary-aware transformer for extracting local details to tackle ambiguous boundaries. Wu \emph{et al.} \cite{wu2022fat} proposed a FAT-Net to extract both local features and long-range dependencies by combining CNNs and transformer while He \emph{et al.} \cite{he2022fully} leveraged sliding windows tokenization to construct hierarchical features. These methods are mainly researched on single-modal skin images, rendering them unsuitable for the fusion of multi-modal skin lesion diagnosis.

Among the studies, Liu \etal \cite{liu2021swin} pointed out there are two main challenges: (1) The resolution of pixels in images is higher than the resolution of words in text, and the calculation cost of vanilla ViT is quadratic to the image resolution. (2) The scale of the target in the image has large variations. CNNs typically use down sampling and overlapping kernels to maximize the utilization of nearby elements at various scales. At the same scale, ViT, which lacks inductive bias, only relies on the self-attention mechanism, leading to worse performance than CNNs with small training dataset. Given the above problems, Liu \etal proposed the Swin Transformer \cite{liu2021swin} with the hierarchical and local priors introduced into the structural design, and the computation cost was reduced. To take advantage of such good characteristics of Swin Transformer, we also introduce it into our framework to replace the deep convolutions as feature extraction backbone, which facilitates the design of our information fusion structure. 

\section{Preliminaries}
In this part, we will briefly describe some preliminary knowledge about Swin Transformer \cite{liu2021swin}, since we utilize it as the feature extraction backbone and the design of the fusion structure in TFormer refers to it. Compared with vanilla ViT \cite{dosovitskiy2020image}, Swin Transformer \cite{liu2021swin} introduces hierarchical and local priors into the structural design to resolve the lack of bias induction. 

\subsection{Patch Merging}
The patch merging operation is critical in achieving the hierarchical structure with the goal of reducing image resolution. As shown in the image feature extraction network in Figure \ref{TFormer}, the first patch merging layer takes $2\times2$ as a step and stitches the values from the same position of adjacent patches (a patch size is $4\times4$, and the dimension of each patch is $4\times4\times3=48$) into a new feature map that is four times smaller than the former. Then the concatenated features are applied to a linear layer with an input dimension of $4C$ and an output of $2C$ to achieve downsampling with a resolution of $2\times$. Through this layer, the resolution is reduced from the original $\frac{H}{4}\times\frac{W}{4}$ to $\frac{H}{8}\times\frac{W}{8}$, and the number of channels is doubled. Compared with pooling, patch merging can utilize more information of the original resolution, which is an indispensable operation for downsampling in the Swin Transformer.

\subsection{Shifted Window Multi-head Self-Attention}

Vanilla ViT mainly depends on global self-attention, which loses locality to a certain extent compared to CNNs \cite{raghu2021vision}. A shifted window multi-head self-attention (SW-MSA) is designed in Swin Transformer. To begin, self-attention is computed within non-overlapping windows, introducing local priors. When each window contains $M\times M$ patches and the entire image contains $h\times w$ patches, the computational complexity of window multi-head self-attention (WMSA) ($\Omega_{WMSA}=4hwC^2+2M^2hwC$) is linear to the image size, effectively lowering the computation cost. Then, after every two layers of transformer blocks, the window of the latter block is shifted by half a window compared with the previous one, so that the information between windows can be connected with modeling ability improved.  Furthermore, the relative position bias is introduced in the calculation of self-attention. The following formula is used to calculate window self-attention (WSA):

\begin{equation}
    WSA(Q,K,V) = SoftMax(\frac{QK^T}{\sqrt{d}}+B)V
\end{equation}
where $Q,K,V\ \in\mathbb{R}^{M^2\times d} $ are the query, key and value sets, $d$ is the key dimension, $M^2$ is the number of patches in a window, and $B \in \mathbb{R}^{M^2\times M^2}$ denotes the position bias. 

\section{Methodology}

\subsection{Overall Architecture of TFormer}

\begin{figure*}
	\centering
	\includegraphics[scale=0.32]{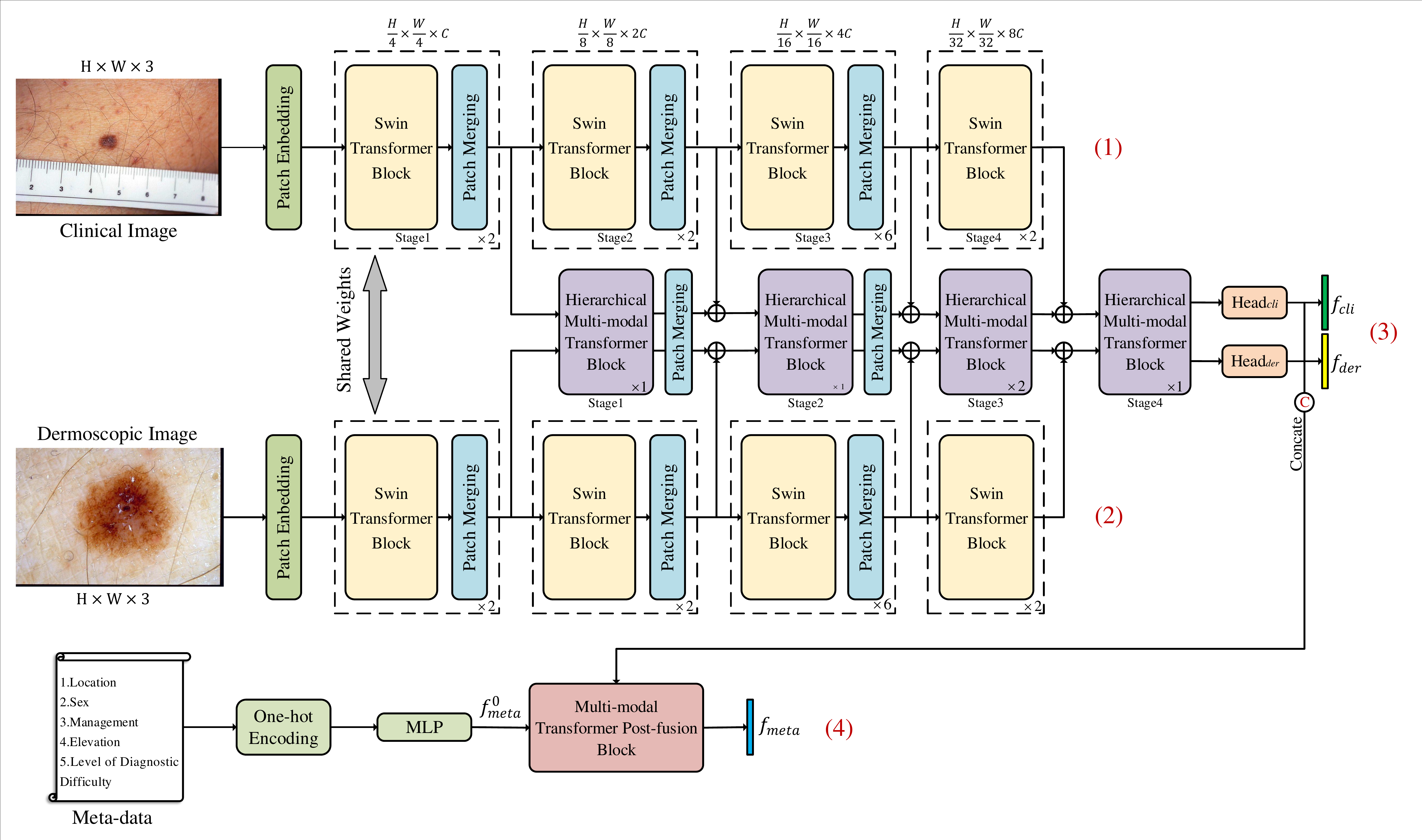}
	\caption{Overall architecture of TFormer: (1) clinical image feature extraction branch (2) dermoscopic image feature extraction branch (3) image modalities fusion branch (for simplification, the dual-branch structure of HMT is represented by a block diagram, and the specific structure can be seen in Figure \ref{HMT+MTP+head}) (4) patients' metadata feature extraction and fusion branch.}
	\label{TFormer}
\end{figure*}

The overall architecture of the proposed TFormer is presented in Figure \ref{TFormer}. With the four-stage Swin Transformer \cite{liu2021swin} as the backbone for image modalities, we can obtain multi-scale features of clinical images and dermoscopic images. For sufficient features fusion between two image modalities, we design a dual-branch hierarchical multi-modal transformer (HMT) block, which is based on cross-attention mechanism, to fuse information across the image modalities. Specifically, each branch in HMT corresponds to a modality (referred to query in self-attention) and fuse the features from the other branch (referred to key and value). To achieve a throughout fusion, we apply a stack of HMT blocks consistent with the stages of the backbone by utilizing hierarchical features form low to high level, and finally obtain the fused feature vectors $f_{cli}$ and $f_{der}$. Previous works often regard image modalities and metadata as equal, which may lead to imbalanced fusion. In stead of fusing three modalities information synchronously, we adopt a "divide and conquer" strategy here. For patients' meta-data modality, we use an MLP to map encoded one-hot vectors to high-dimensional space, and then we propose a multi-modal transformer post-fusion (MTP) block which can calculate the cross-attention among the extracted features of the image modalities and that of meta-data, leading to the fused feature vector $f_{meta}$. In the end, we concatenate the above three extracted vectors and pass them into a classification layer to obtain the final diagnosis results. More details are described in the following.

\subsection{Hierarchical Multi-modal Transformer Block}

\begin{figure*}
	\centering
	\includegraphics[scale=0.45]{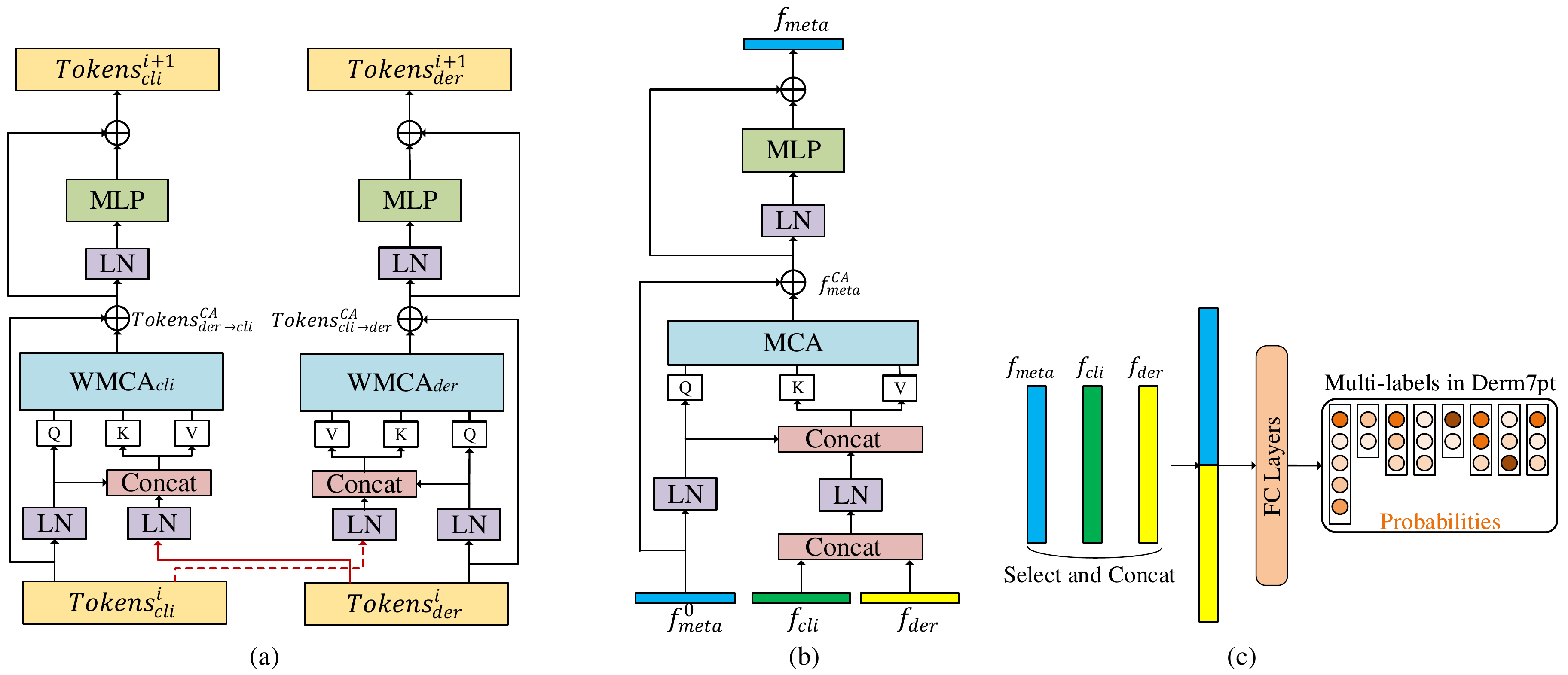}
	\caption{(a) Hierarchical multi-modal transformer (HMT) block. (b) Multi-modal transformer post-fusion (MTP) block. (c) The structure of multi-label classification layer.}
	\label{HMT+MTP+head}
\end{figure*}

The specific structure of the HMT block is shown in Figure \ref{HMT+MTP+head} (a). The two branches respectively focus on dermoscopic images and clinical images, while effectively modeling the commonality and complementarity between modalities. Based on WSA, we design a window multi-head cross-attention (WMCA). For the sake of brevity, we will describe how to pass information from one modality to the other modality with the dermoscopic branch (the right branch in Figure \ref{HMT+MTP+head} (a)) as an example. Assuming that the dermoscopic image features of stage $i$ are ${Tokens}^i_{der} \in\mathbb{R}^{{M}^2\times C}$, and the clinical image features are ${Tokens}^i_{cli} \in\mathbb{R}^{{M}^2\times C}$, we take ${Tokens}^i_{der}$ as the query ($Q^\prime$) of WMCA and the concatenation of ${Tokens}^i_{der}$ and ${Tokens}^i_{cli}$ as the key ($K^\prime$) and value ($V^\prime$). As shown in the following formula:

\begin{equation}
    Q^\prime=LN({{Tokens}^i}_{der})W^{Q^\prime}
\end{equation}
\begin{equation}
    K^\prime=Concat(LN({{Tokens}^i}_{der}),LN({{Tokens}^i}_{cli}))W^{K^\prime}
\end{equation}
\begin{equation}
    V^\prime=Concat(LN({{Tokens}^i}_{der}),LN({{Tokens}^i}_{cli}))W^{V^\prime}
\end{equation}
where $Q^\prime\in\mathbb{R}^{M^2\times C}$, $K^\prime$ and $V^\prime\in\mathbb{R}^{{2M}^2\times C}$, $W^{Q^\prime}$, $W^{K^\prime}$, and $W^{V^\prime}$ are linear mapping parameter metrics, while $LN$ denotes the layer norm operation. Then, the WMCA from clinical branch to dermoscopic branch can be calculated as following:

\begin{equation}
    {\rm WMCA}_{cli\rightarrow d e r}(Q^\prime,K^\prime,V^\prime)=SoftMax(\frac{Q^\prime{K^\prime}^T}{\sqrt{C/ h}}+B^\prime)V^\prime
\end{equation}
where $B^\prime\epsilon\mathbb{R}^{M^2\times{2M}^2}$ is the relative position bias, $M^2$ is the number of patches in a window and $h$ denotes the number of heads.

The output obtained through multi-head attention WMCA is connected to the input ${{Tokens}^i}_{der}$ in a residual operation, and then the further obtained output ${Tokens}_{cli\rightarrow der}^{CA}$ is input into an MLP forward propagation network with a LN and a residual shortcut.  Finally, the fused features of dermoscopic image that integrate clinical image information are obtained by:

\begin{equation}
    {Tokens}_{cli\rightarrow der}^{CA}={\rm WMCA}_{cli\rightarrow der}(Q^\prime,K^\prime,V^\prime)+{{Tokens}^i}_{der}
\end{equation}

\begin{equation}
    {Tokens}_{der}^{i+1}=MLP(LN({Tokens}_{cli\rightarrow der}^{CA}))+{Tokens}_{cli\rightarrow der}^{CA}
\end{equation}

The hidden representations of the dermoscopic image are then guided by clinical image with the above single cross-attention branch, which can fuse the different modalities effectively in a latent adaptation. Similarly, another branch is used for the flow of dermoscopic image information to the clinical image. Therefore, the design of the dual-branch structure is conducive to the bimodal interaction.

We apply a stack of HMT blocks to fuse the features throughout the network to fully utilize the hierarchical features with different scales extracted by the backbone. To improve the ability of modeling inter-modality dynamics under the variable scale, we leverage both the features extracted by the backbone and the output of the previous HMT block as the input of the current HMT block, which can also bring the feature information in the network better liquidity. The final output tokens at the last stage are fed into two $head$ modules composed of a pooling operation and a fully connected layer to get the fused feature vectors $f_{cli}$ and $f_{der}$, which can be formulated as:
\begin{equation}
    f_{cli}={head}_{cli}({Tokens}_{cli}^N)
\end{equation}

\begin{equation}
    f_{der}={head}_{der}({Tokens}_{der}^N)
\end{equation}

\subsection{Multi-modal Transformer Post-fusion Block}

We input the meta-data into an MLP (containing two FC layers) after one-hot encoding, to obtain the metadata feature $f_{meta}^0$. Then, $f_{meta}^0$ and the fused features $f_{cli}$ and $f_{der}$ are fed into the designed MTP block to pass the image modality information to the text modality. As illustrated in Figure \ref{HMT+MTP+head} (b), MTP is a single-branch multi-modal transformer with the purpose of getting the fused meta-data feature under the guidance of image information. To be specific, taking $f_{meta}^0$ as the query, $f_{meta}^0$, the concatenation of $f_{cli}$ and $f_{der}$ as key and value, we then have:

\begin{equation}
    \widetilde{Q}=LN(f_{meta}^0)W^{\widetilde{Q}}
\end{equation}

\begin{equation}
    \widetilde{K}=Concat(LN(f_{meta}^0),LN(Concat(f_{cli},f_{der})))W^{\widetilde{K}}
\end{equation}

\begin{equation}
    \widetilde{V}=Concat(LN(f_{meta}^0),LN(Concat(f_{cli},f_{der})))W^{\widetilde{V}}
\end{equation}

Then, leveraging the multi-head cross-attention (MCA) mechanism, we can get the output $f_{meta}^{CA}$ based on $\widetilde{Q},\widetilde{K},\widetilde{V}$ and feed the output into an MLP layer with a residual shortcut. Finally, we get the fused feature $f_{meta}$. The process can be formulated as:

\begin{equation}
    f_{meta}^{CA}={MCA (\widetilde{Q},\widetilde{K},\widetilde{V})+f}_{meta}^0
\end{equation}

\begin{equation}
    f_{meta}=MLP(LN(f_{meta}^{CA}))+f_{meta}^{CA}
\end{equation}

\subsection{Multi-label Classification Layer}

Throughout the modules above, we get three feature vectors that integrate other modal information $f_{meta}$, $f_ {cli}$, $f_ {der}$. As shown in Figure \ref{HMT+MTP+head}(c), these vectors are selectively used and concatenated, then are input into the multi-label classification layer composed of several fully connection (FC) layers to obtain the probabilities of various labels for the diagnosis of skin diseases. We here make use of $f_{meta}$ and $f_{der}$, which perform better results suggested by experiments in Section \ref{select features}. 

\subsection{Loss}
Following \cite{bi2020multi,tang2022fusionm4net}, we simply use the cross-entropy loss to evaluate the multi-label predictions. The formula is as follows:

\begin{equation}
   Loss=\frac{1}{N}\sum_{i=1}^{N}{CE(p_i,y_i)}
\end{equation}
where $\mathbf{y}$ means the ground truth of the multi-labels, $p_{i}$a nd $y_{i}$ are the predicted value and true value of the i-th label respectively, $N$ represents the number of multi-labels (8 in this study), and $CE$ represents the cross-entropy loss.

\section{Experiments} \label{Experiments}
In this section, we first introduce the dataset, evaluation metrics, and implementation details. For the experiment results, we first show the comparisons of single- and multi-modal network performance. The ablation study on the structure of TFormer is then conducted and analyzed. Finally, we compare our method with the current state-of-the-art methods.
\subsection{Dataset}
\begin{table}[!t]
\caption{\label{Derm7pt}Details of the Derm7pt dataset.}
\renewcommand{\arraystretch}{1.1}
\begin{tabular}{llll}
\hline \hline
Label                                                                                  & name                 & abbrev & Num \\ \hline  \hline
\multirow{5}{*}{Diagnosis (DIAG)}                                                      & Melanoma             & MEL    & 252 \\
                                                                                       & Nevus                & NEV    & 575 \\
                                                                                       & Seborrheic Keratosis & SK     & 45  \\
                                                                                       & Basal Cell Carcinoma & BCC    & 42  \\
                                                                                       & Miscellaneous        & MISC   & 97  \\ \hline
\multicolumn{4}{c}{Seven Point Criteria}                                                                                     \\ \hline
\multirow{3}{*}{\begin{tabular}[c]{@{}l@{}}Pigment \\ Network (PN)\end{tabular}}        & Absent               & ABS    & 400 \\
                                                                                       & Typical              & TYP    & 381 \\
                                                                                       & Atypical             & ATP    & 230 \\ \hline
\multirow{2}{*}{\begin{tabular}[c]{@{}l@{}}Blue Whitish \\ Veil (BWV)\end{tabular}}    & Absent               & ABS    & 816 \\
                                                                                       & Present              & PRS    & 195 \\ \hline
\multirow{3}{*}{\begin{tabular}[c]{@{}l@{}}Vascular \\ Structures (VS)\end{tabular}}   & Absent               & ABS    & 823 \\
                                                                                       & Regular              & REG    & 117 \\
                                                                                       & Irregular            & IR     & 71  \\ \hline
\multirow{3}{*}{Pigmentation (PIG)}                                                    & Absent               & ABS    & 588 \\
                                                                                       & Regular              & REG    & 118 \\
                                                                                       & Irregular            & IR     & 305 \\ \hline
\multirow{3}{*}{Streaks (STR)}                                                         & Absent               & ABS    & 653 \\
                                                                                       & Regular              & REG    & 107 \\
                                                                                       & Irregular            & IR     & 251 \\ \hline
\multirow{3}{*}{\begin{tabular}[c]{@{}l@{}}Dots and \\ Globules (DaG)\end{tabular}}    & Absent               & ABS    & 229 \\
                                                                                       & Regular              & REG    & 334 \\
                                                                                       & Irregular            & IR     & 448 \\ \hline
\multirow{2}{*}{\begin{tabular}[c]{@{}l@{}}Regression \\ Structures (RS)\end{tabular}} & Absent               & ABS    & 758 \\
                                                                                       & Present              & PRS    & 253 \\ \hline \hline
\end{tabular}
\end{table}

\footnotetext[1]{7-point checklist is a rule-based diagnostic algorithm proposed in \cite{argenziano1998epiluminescence}. It takes multiple dermoscopy criteria into account and can help dermatologists distinguish between benign and malignant skin tumors.}

We use a publicly available multi-modal dataset Derm7pt \cite{kawahara2018seven} to evaluate our method, which provide three modalities for skin lesion classification task with eight types of labels (diagnostic label and 7-point checklist labels \footnotemark[1]), thus it can be summarized as a multi-label multi-classification task. There are 1011 cases in total, each case contains a dermoscopic image, a clinical image and patient meta-data with information such as location, gender, management, elevation, and diagnostic difficulty level. These 1011 cases are officially divided into three sets including 413 training cases, 203 validation cases, and 395 test cases. The 7-point checklist labels include: pigment network (PN), pigmentation (PIG) blue whitish veil (BWV), vascular structures (VS), streaks (STR), dots and globules (DaG), and regression structures (RS). Table \ref{Derm7pt} shows the diagnostic and the 7-point checklist labels in detail.

\subsection{Evaluation Metrics}
We use average accuracy (avg.) as the main evaluation metric for comparison, other metrics including sensitivity (SEN), specificity (SPE), precision (PRE), F-score and the averages of all labels are used to compare with other state-of-the-art methods. The calculation formulas are as follows:

\begin{equation}\label{Accuracy}
Accuracy = \dfrac{TP+TN}{TP+TN+FP+FN}
\end{equation}

\begin{equation}\label{Precision}
Precision = \dfrac{TP}{TP+FP}
\end{equation}

\begin{equation}\label{Sensitivity}
Sensitivity = TPR = \frac{TP}{TP+FN}
\end{equation}

\begin{equation}\label{Specificity}
Specificity = 1-FPR = 1-\dfrac{FP}{FP+TN} = \dfrac{TN}{FP+TN}
\end{equation}

\begin{equation}
    F-score = (1+\beta^2)\dfrac{Precision \times Sensitivity}{\beta^2Precision+Sensitivity}
\end{equation}
where $TPR$ means true positive rate, $FPR$ means false positive rate. $TP$, $FN$, $TN$ and $FP$ represent the number of true positive, false negative, true negative and false positive, respectively, and $\beta$ is set to 1 (F1-score).

\subsection{Implement Details}
The algorithms in this study are all written in Pytorch and run on a PC with an NVIDIA A100 GPU.  During training, the batch size is set to 32. We use Adam \cite{kingma2014adam} as the optimizer with the initial learning rate of 1e-4 and the weight decay of 1e-4. The cosine annealing strategy \cite{loshchilov2016sgdr} is used for the learning rate decay, 100 epochs are trained, and the model with the highest avg. on the validation set is saved for testing. Considering the small dataset, we chose Swin-Tiny as the feature extraction network and used weights pre-trained on ImageNet-1K \cite{deng2009imagenet}. The input image size of network is $224\times224\times3$ ($H\times W\times3$ in Figure \ref{architecture}), and the length of the patient meta-data after one-hot encoding is 20. Data augmentation methods include random vertical and horizontal flips, shifts, rotation, and brightness contrast conversion, which is same as \cite{tang2022fusionm4net}. The results of the experiments are averaged multiple times and the standard deviations (SD) are reported. The dimensions of the FC layers in both $head$ and MLP are set to 128. Furthermore, unless otherwise specified, the number of HMT blocks used in each stage in TFormer is 1. Since the same dataset is used, for the comparison methods, the experimental parameters and results are from their respective papers.

\subsection{Comparisons Between Single- and Multi-modality leaning}
We first conduct experiments by using single- or multi-modality data. Furthermore, for multi-modality learning we also compare our TFormer with the method illustrated with the solid line in Figure \ref{architecture} (a), which simply use concatenation (Concat.) for modalities fusion\cite{kawahara2018seven,lu2021multimodal}. The results are reported in Table \ref{Single_Multi}, in which the hook means that the corresponding modality is used to classify lesions. Table \ref{Single_Multi} shows that our TFormers using multi-modality data all outperform the method using one modality, while the concatenation method does not. For example, the concatenation method of fusing dermoscopic and clinical modalities has a lower avg. (74.67\%) than the results (74.87\%) only using dermoscopic images, indicating that the late-fusion strategy leads to an insufficient utilization of inter-modality complementarity. This may be due to the simple concatenation interfere with the network's mining in texture details (such as vascular structures and pigment networks in dermoscopic images), although clinical images can introduce global color and shape information. In comparison, TFormer can use hierarchical features at different stages to guide the effective interaction of intra-modality and inter-modality information from shallow layers to deep ones. Finally our method achieves the best performance (avg. 76.62\%) when using all three modalities, demonstrating its superior ability in feature integration.


\begin{table*}
\caption{\label{Single_Multi} Comparisons for single- and multi-modality skin lesion classification (accuracy \%).  }
\renewcommand{\arraystretch}{1.3}
\begin{threeparttable}
\resizebox{\textwidth}{!}{
\begin{tabular}{c|ccc|cccccccc|c}
\hline\hline
     & Derm. & Cli. & Meta & DIAG  & PN  & BMV  & VS& PIG  & STR   & DaG   & RS & {\color[HTML]{CB0000} avg.}          \\ \hline\hline
 & \Checkmark    &     &      & 72.07$\pm$1.45          & 70.29$\pm$1.38          & 88.43$\pm$0.65          & 80.42$\pm$1.80         & 69.80$\pm$2.10         & 75.70$\pm$1.24          & 63.80$\pm$2.10         & 78.79$\pm$1.91          & 74.87$\pm$0.39         \\
 &     & \Checkmark    &      & 64.98$\pm$0.73          & 60.68$\pm$2.27         & 83.46$\pm$2.03          & 81.18$\pm$0.43          & 61.10$\pm$0.52          & 66.67$\pm$2.50          & 55.19$\pm$1.62          & 75.11$\pm$1.04          & 68.54$\pm$0.13          \\
\multirow{-3}{*}{\begin{tabular}[c]{@{}c@{}}Single-\\ Modality\end{tabular}} &     &     & \Checkmark     & 69.96$\pm$1.02          & 58.31$\pm$0.52          & 84.73$\pm$1.06          & 79.24$\pm$0.05          & 58.23$\pm$0.21          & 71.40$\pm$0.83          & 60.59$\pm$0.66          & 71.73$\pm$0.31          & 69.27$\pm$0.17          \\ \hline
 &  \Checkmark   &\Checkmark     &      & 71.39$\pm$1.03          & 70.04$\pm$0.73          & 87.93$\pm$1.38          & 82.53$\pm$0.71          & 69.79$\pm$1.25          & 73.84$\pm$0.63          & 61.69$\pm$1.19          & 80.17$\pm$1.72          & 74.67$\pm$0.61          \\
& \Checkmark    &     &\Checkmark      & 73.08$\pm$1.04          & 68.10$\pm$2.52          & 87.68$\pm$0.78          & 79.85$\pm$1.83          & 70.29$\pm$1.33          & 76.11$\pm$1.45          & 65.74$\pm$1.25          & 78.48$\pm$1.62          & 74.92$\pm$0.71          \\
 &     &\Checkmark     & \Checkmark     & 67.26$\pm$0.66          & 61.94$\pm$1.06          & 84.31$\pm$2.28          & 80.17$\pm$0.78          & 63.97$\pm$2.81          & 68.52$\pm$0.73          & 59.32$\pm$0.72          & 74.68$\pm$1.99          & 70.02$\pm$0.37          \\
\multirow{-4}{*}{Concat.}  & \Checkmark    &\Checkmark     &\Checkmark      & 71.06$\pm$1.21          & 69.79$\pm$0.63          & 87.42$\pm$0.12          & 82.45$\pm$1.58          & 70.55$\pm$0.86          & 72.83$\pm$0.52          & 62.49$\pm$0.96          & 79.49$\pm$0.91          & 74.51$\pm$0.12          \\ \hline
& \Checkmark    & \Checkmark    &      & 73.00$\pm$0.86          & 70.89$\pm$1.80          & 86.42$\pm$0.48          & \textbf{83.45$\pm$0.63} & 68.77$\pm$1.55          & 74.01$\pm$1.17          & 64.90$\pm$0.43          & 81.26$\pm$0.36          & 75.34$\pm$0.30          \\
 & \Checkmark    &     & \Checkmark     & 75.95$\pm$0.71          & 70.81$\pm$0.86          & 87.43$\pm$0.20          & 79.91$\pm$1.38          & \textbf{70.72$\pm$0.93} & 76.88$\pm$0.32          & \textbf{65.06$\pm$1.43} & 78.48$\pm$0.90          & 75.64$\pm$0.06          \\
&     & \Checkmark    & \Checkmark     & 73.59$\pm$0.43          & 64.14$\pm$0.78          & 84.05$\pm$1.35          & 80.59$\pm$1.21          & 61.18$\pm$0.52   & 69.20$\pm$0.95          & 59.58$\pm$2.07          & 74.85$\pm$1.92          & 70.90$\pm$0.38          \\
\multirow{-4}{*}{\begin{tabular}[c]{@{}c@{}}TFormer\\ (Ours)\end{tabular}}   &  \Checkmark   & \Checkmark    & \Checkmark     & \textbf{78.48$\pm$0.35} & \textbf{73.08$\pm$0.32} & \textbf{88.13$\pm$0.21} & 81.77$\pm$1.09          & 69.37$\pm$1.15          & \textbf{76.03$\pm$0.98} & 64.64$\pm$0.52          & \textbf{81.44$\pm$0.83} & \textbf{76.62$\pm$0.11} \\ \hline\hline
\end{tabular}}
\begin{tablenotes}
\footnotesize
\item Derm. is the abbreviation of dermoscopic image, Cli. is the abbreviation of clinical image and Meta means meta-data. 
\item Concat. and TFormer are all multi-modality learning methods.
\end{tablenotes}
\end{threeparttable}
\end{table*}

\subsection{Controlled Experiments}
In this part, we first conduct experiments and analyses to verify the effectiveness of the structure design of TFormer. Then, we discuss the effect of different fused features selection strategies in the classification layer.
\subsubsection{Different Structure Design of TFormer} \label{ablation}

\textbf{Dual Branch in HMT.} Table \ref{HMT} presents the results of the ablation study of HMT. The results show that the fused features of clinical images generated under the guidance of dermoscopic images have better data representative ability than the guidance of clinical images. However, the separate branch is biased towards a single flow from one modality to another, and the results are worse than dual-branch. Thus, the effectiveness of the dual-branch fusion structure can be demonstrated, which enhances the information interaction in the shallow-to-deep inference process of the network.
\begin{table*}
\renewcommand{\arraystretch}{1.3}
\caption{\label{HMT} The results of ablation study for HMT  (accuracy \%).  }
\begin{threeparttable}
\resizebox{\textwidth}{!}{
\begin{tabular}{c|cccccccc|c}
\hline \hline
Branch     & DIAG                                 & PN                                   & BMV                                  & VS                                   & PIG                                  & STR                                  & DaG                                  & RS                                   & {\color[HTML]{CB0000} avg.}          \\ \hline\hline
Cli.$\rightarrow$ Derm. & 73.07$\pm$0.52          & 69.95$\pm$0.48          & 85.91$\pm$0.12          & 81.94$\pm$0.67          & 68.51$\pm$0.52          & 73.42$\pm$1.29          & 60.75$\pm$0.34          & 80.34$\pm$0.73          & 74.24$\pm$0.16          \\
Derm.$\rightarrow$ Cli. & \textbf{73.25$\pm$1.26} & 69.87$\pm$1.36          & \textbf{86.67$\pm$1.18} & 83.06$\pm$0.04          & 68.35$\pm$1.07          & \textbf{74.51$\pm$0.98} & 62.03$\pm$0.95          & 80.34$\pm$0.78          & 74.76$\pm$0.25          \\
Dual       & 73.00$\pm$0.86          & \textbf{70.89$\pm$1.80} & 86.42$\pm$0.48          & \textbf{83.45$\pm$0.63} & \textbf{68.77$\pm$1.55} & 74.01$\pm$1.17          & \textbf{64.90$\pm$0.43} & \textbf{81.26$\pm$0.36} & \textbf{75.34$\pm$0.30} \\ \hline\hline
\end{tabular}}
\begin{tablenotes}
\footnotesize
\item Cli.$\rightarrow$ Derm. denotes the right branch in Figure \ref{HMT+MTP+head} (a) and Derm.$\rightarrow$ Cli. denotes the left one.
\end{tablenotes}
\end{threeparttable}
\end{table*}

\textbf{Ablation Study for Hierarchical Features Choice.} There are HMT blocks between different stages of feature extractors in Figure \ref{TFormer} to fuse the two modalities. To verify the effectiveness of feature fusion at different levels, we conduct experiments by continuously reducing HMT blocks from shallow to deep. The results in Table \ref{DIWS} (the first four rows) show that with the number of HMT blocks reduced, the accuracy of network decreases, which indicate that features at different levels are all important in modeling inter-modality dynamics. Shallow features focus on the local structures of lesions between modalities, while deep features pay more attention to semantic information. Furthermore, variant scales of stages can better adapt to the unaligned spatial resolution in dermoscopic and clinical images. At the same time, we can see that when remove the HMT in the stage 1, the avg. decreases even more, suggesting that the shallow information can retain more information from the original image, which contains more local details and serves as the foundation for extracting high-level semantic features.

\begin{figure*}
	\centering
	\includegraphics[scale=0.65]{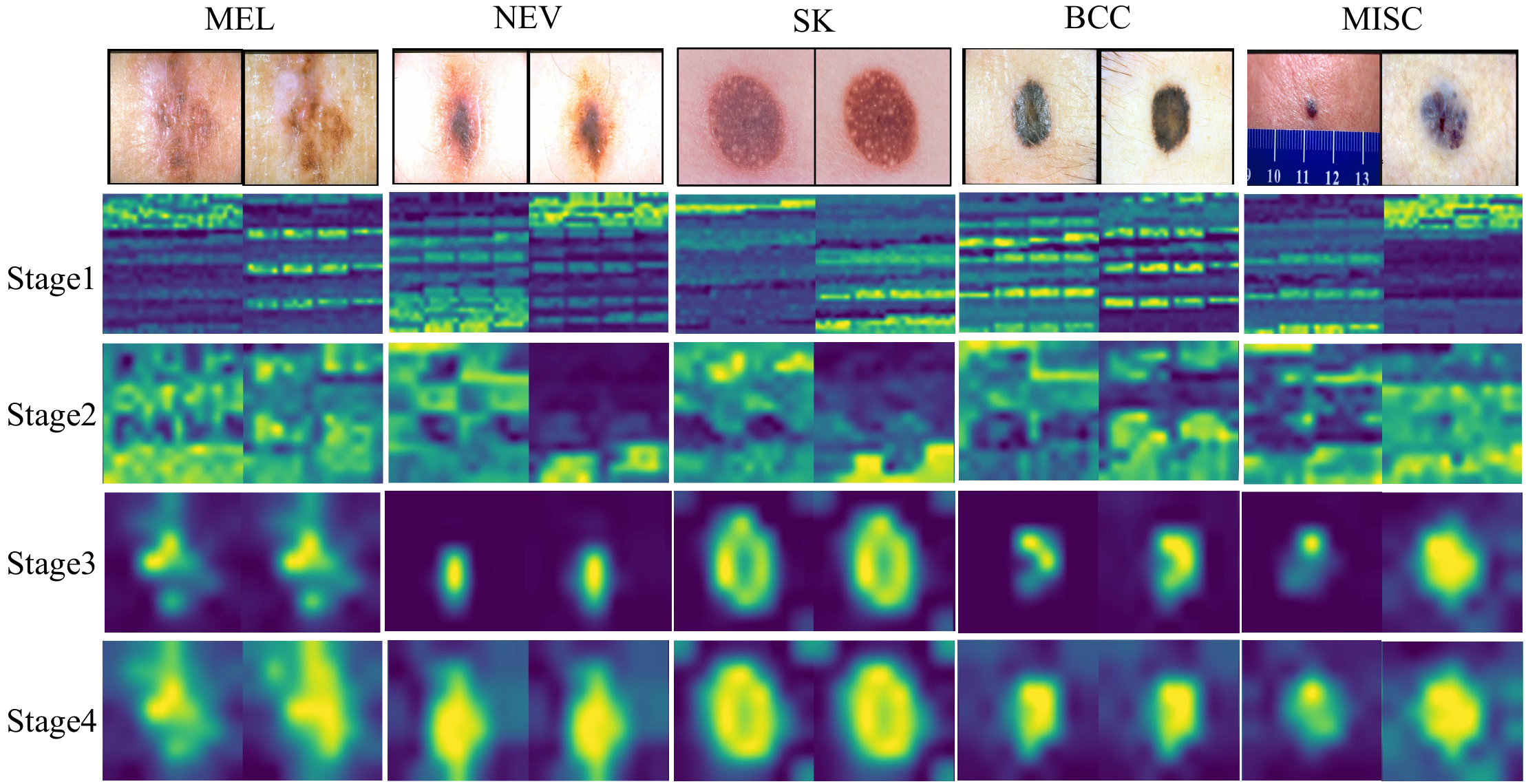}
	\caption{Attention behavior. The left side of each input group is a clinical image, and the right side is a dermoscopic image. Window attentions in the low stages of network focus on local details while high-level attentions focus globally.}
	\label{attention_maps}
\end{figure*}

To explore how the multi-head cross-attention works in the HMT block, we visualized the attention maps of HMT in different stages, and randomly selected one from the attention maps of two branches as shown in Figure \ref{attention_maps}. It can be seen that the window attentions in shallow layers focus on local details and texture features. As the depth of the network increases, the receptive field expands and the attentions become more global, causing the network to concentrate on the area with skin lesions.

\textbf{Depth Increase and Weight Sharing.} In previous experiments, the depth of the HMT block is 1 for each stage. In this section, we try to deepen the HMT to see if a deeper fusion module can improve the ability of learning cross-modality features. The results in Table \ref{DIWS} show that when the depth increases from shallow stages to deep ones, the avg. first increases and then decreases. This could be attributed to the higher increased amounts of parameters when the same number of HMT are added as the depth of the feature extraction network increases. Due to the small sample size of Derm7pt (training samples $< 500$), the network is more prone to overfitting. To alleviate the phenomenon, we then try to use weight sharing strategy for the feature extractors. The results show that, when the number of parameters decreases by sharing the weights, avg. of the networks are all improved. Furthermore, the weight-sharing networks can jointly learn features of dermoscopic and clinical images, improving the ability to model the inter-modality consistency and complementarity.
Ultimately, we find that TFormer achieves a higher avg. (average of 77.40\% and highest of 77.99\%) when increasing the HMT in stage 3. One of the potential reasons is that high-level features have a stronger data representative ability than shallow-level ones, as well as the cross-modalities interaction. However, when the transformer becomes much deeper (such as stage 4), the multi-heads in this level basically focus on long-distance features \cite{raghu2021vision} with a lack of local details, leading to a harder information flow between modalities.

\begin{table*}[!t]
\renewcommand{\arraystretch}{1.3}
\caption{\label{DIWS} The results of the study for selecting hierarchical features, and the study of depth increase and weight sharing (accuracy \%). }
\resizebox{\textwidth}{!}{
\begin{tabular}{cccc|c|cccccccc|c}
\hline\hline
\multicolumn{4}{c|}{Stages} &\multirow{2}{*}{\begin{tabular}[c]{@{}c@{}}Shared\\ Weights\end{tabular}} & \multirow{2}{*}{DIAG} & \multirow{2}{*}{PN} &\multirow{2}{*}{BMV}&\multirow{2}{*}{VS}&\multirow{2}{*}{PIG}&\multirow{2}{*}{STR}&\multirow{2}{*}{DaG}&\multirow{2}{*}{RS}&\multirow{2}{*}{{\color[HTML]{CB0000} avg.}} \\\cline{1-4}         
1 & 2 & 3 & 4           &                  &                         &                           &                          &                           &              &               &               &               &  \\ \hline\hline

-&-&-&1                   &   \ding{55}            & 77.32$\pm$1.25 & 70.89$\pm$1.26 & 87.25$\pm$0.48 & 81.69$\pm$0.52 & 70.38$\pm$1.45          & 75.22$\pm$0.92          & 65.76$\pm$2.17          & 80.09$\pm$0.60          & 76.07$\pm$0.41                     \\

-&-&1&1                   &   \ding{55}            & 77.55$\pm$0.96                     & 73.00$\pm$1.26                     & 86.75$\pm$0.83                     & 81.77$\pm$0.21                     & 68.78$\pm$0.66          & \textbf{76.88$\pm$0.83} & 64.39$\pm$0.73          & 81.18$\pm$1.06          & 76.29$\pm$0.21                     \\
-&1&1&1                       &   \ding{55} &  \textbf{78.82$\pm$1.41}            & 71.56$\pm$0.83                     & 87.93$\pm$0.32                     & \textbf{82.45$\pm$0.96}            & 69.03$\pm$1.14          & 74.68$\pm$1.15          & \textbf{66.25$\pm$1.02} & 80.00$\pm$0.42          & 76.34$\pm$0.30       \\

1&1&1&1                   &   \ding{55}            & 78.48$\pm$0.35          & 73.08$\pm$0.32          & \textbf{88.13$\pm$0.21} & 81.77$\pm$1.09          & 69.37$\pm$1.15          & 76.03$\pm$0.98          & 64.64$\pm$0.52                     & 81.44$\pm$0.83          & 76.62$\pm$0.11          \\ \hline
       & & &                         &    \ding{55}          & 77.72$\pm$0.21          & 73.08$\pm$0.43          & 86.42$\pm$0.66          & 82.53$\pm$0.95          & 70.13$\pm$0.54          & \textbf{77.21$\pm$0.72} & 67.76$\pm$1.50                          & 80.83$\pm$0.77          & 76.92$\pm$0.13          \\
\multirow{-2}{*}{2}&\multirow{-2}{*}{1} &\multirow{-2}{*}{1} &\multirow{-2}{*}{1} &   \Checkmark            & 78.06$\pm$1.76          & 71.82$\pm$0.63          & 87.43$\pm$0.32          & 82.19$\pm$1.14          & 70.21$\pm$0.63          & 76.87$\pm$0.12          & \textbf{67.91$\pm$0.41} & 81.60$\pm$0.83          & 77.01$\pm$0.26          \\ \hline
   & & &                             &       \ding{55}        & 77.56$\pm$1.06          & 72.15$\pm$0.62          & 86.75$\pm$0.52          & \textbf{83.37$\pm$0.67} & 69.62$\pm$1.07          & 74.68$\pm$1.64          & 65.82$\pm$0.90                          & 80.84$\pm$1.17          & 76.35$\pm$0.29          \\
\multirow{-2}{*}{1}&\multirow{-2}{*}{2}&\multirow{-2}{*}{1}&\multirow{-2}{*}{1} &    \Checkmark          & 78.56$\pm$0.42          & 72.06$\pm$0.64          & 87.51$\pm$0.84          & 82.19$\pm$0.67          & \textbf{70.63$\pm$1.09} & 75.11$\pm$1.21          & 67.26$\pm$1.14                          & 81.52$\pm$0.55          & 76.85$\pm$0.13          \\ \hline
    & & &                            &     \ding{55}         & 77.38$\pm$0.52          & 72.74$\pm$1.76          & 86.16$\pm$0.79          & 82.19$\pm$1.06          & 68.69$\pm$1.17          & 75.61$\pm$0.67          & 66.08$\pm$1.26                          & 80.59$\pm$0.32          & 76.18$\pm$0.18          \\
\multirow{-2}{*}{1}&\multirow{-2}{*}{1}&\multirow{-2}{*}{2}&\multirow{-2}{*}{1} &  \Checkmark             & \textbf{79.49$\pm$0.54} & \textbf{74.34$\pm$0.93} & 86.67$\pm$0.95          & 83.03$\pm$0.55          & 70.29$\pm$1.67          & 76.71$\pm$1.64          & 66.85$\pm$0.75                          & \textbf{82.11$\pm$1.45} & \textbf{77.40$\pm$0.59} \\ \hline
       & & &                         &    \ding{55}           & 78.06$\pm$1.06          & 72.74$\pm$1.14          & 87.09$\pm$1.61          & 81.60$\pm$0.73          & 68.44$\pm$0.98          & 74.51$\pm$0.98          & 65.82$\pm$0.82                          & 81.03$\pm$0.57          & 76.16$\pm$0.05          \\
\multirow{-2}{*}{1}&\multirow{-2}{*}{1}&\multirow{-2}{*}{1}&\multirow{-2}{*}{2} &   \Checkmark           & 78.14$\pm$0.43          & 71.82$\pm$1.02          & 87.76$\pm$0.32          & 82.70$\pm$0.12          & 69.70$\pm$0.48          & 76.62$\pm$1.56          & 65.49$\pm$0.43                          & 81.05$\pm$1.29          & 76.66$\pm$0.16          \\ \hline\hline
\end{tabular}}
\begin{tablenotes}
\footnotesize
\item - in the columns of ``Stages" means the HMT block is removed in the corresponding stage.
\end{tablenotes}
\end{table*}

\subsubsection{Different Fused Features Selection in Classification Layer}\label{select features}

\begin{table*}
\renewcommand{\arraystretch}{1.3}
\caption{\label{FS} The results of fused features selection in classification layer (accuracy \%). }
\resizebox{\textwidth}{!}{
\begin{tabular}{ccc|cccccccc|c}
\hline\hline
$f_{cli}$ & $f_{der}$ & $f_{meta}$ & DIAG                        & PN                          & BMV                                  & VS                                   & PIG                         & STR                                  & DaG                                  & RS                          & {\color[HTML]{CB0000} avg.} \\ \hline\hline
          &           &  \Checkmark          & 77.81$\pm$0.52 & \textbf{73.76$\pm$1.58} & 85.82$\pm$0.55 & 81.69$\pm$0.48          & 69.79$\pm$1.26 & 74.01$\pm$1.34          & 66.24$\pm$2.03          & 80.76$\pm$0.72 & 76.23$\pm$0.14 \\
      \Checkmark    &           &    \Checkmark        & 77.56$\pm$0.96 & 73.17$\pm$1.65 & \textbf{88.44$\pm$0.52}          & 81.37$\pm$0.66          & 68.86$\pm$1.29 & 73.84$\pm$1.66          & 64.98$\pm$0.48 & 80.93$\pm$1.58 & 76.14$\pm$0.54 \\
         &  \Checkmark          &    \Checkmark        & \textbf{78.48$\pm$0.35} & 73.08$\pm$0.32 & 88.13$\pm$0.21          & 81.77$\pm$1.09          & \textbf{69.37$\pm$1.15} & \textbf{76.03$\pm$0.98} & 64.64$\pm$0.52          & \textbf{81.44$\pm$0.83} & \textbf{76.62$\pm$0.11} \\
      \Checkmark    &   \Checkmark        &  \Checkmark          & 77.05$\pm$0.43 & 72.06$\pm$0.98 & 86.83$\pm$0.55          & \textbf{82.45$\pm$0.78} & 68.69$\pm$0.32 & 74.48$\pm$0.68          & \textbf{66.24$\pm$1.67}          & 80.59$\pm$0.86 & 76.05$\pm$0.19 \\ \hline\hline
\end{tabular}}
\end{table*}

The clinical image features $f_{cli}$ (fused with dermoscopic images), the dermoscopic image features $f_{der}$ (fused with clinical images) and the meta-data features $f_{meta}$ (fused with dermoscpic and clinical modalities) are obtained via the network structure in Figure \ref{TFormer}. Then, as shown in Figure \ref{HMT+MTP+head}(3), we select and concatenate these fused features into the classification layer. We define the selection criterion as that the concatenated features should contain the information from three modalities (thus excluding the use of only $f_{der}$ or $f_{cli}$). The results in Table \ref{FS} show that when using $f_{der}$ and $f_{meta}$, the avg. is higher and the SD is smaller. Thus we find that the accuracy is not positively correlated with the number of fused features involved in the final decision, which can result form redundancy and confusion between the fused features $f_{der}$, $f_{cli}$ and $f_{meta}$. We can see that the participation of $f_{cli}$ in classification will bring about a lower avg. (for example, the avg. of using all three features in Table \ref{FS} is 0.49\% lower than that of using $f_{der}$ and $f_{meta}$). The reason for this could be that the features $f_{cli}$ have introduced too many surface characteristics such as shape and geometry in clinical images. Nevertheless, in routine clinical diagnosis, these global features contribute less to the classification of the seven-point checklist labels (doctors can obtain more subtle structures including blood vessels, dots, and spheres etc. beneath the skin surface from dermoscopic images, which are invisible to the naked eyes), and this portion of the contributed information has been implicitly passed to the fused $f_{der}$ via HMT blocks. Besides, it is noteworthy that the importance of the clinical image branch in HMT does not be diminished. On the contrary, it is with this branch that the network can better pass the complementary intrinsic features of clinical images to the fused features of dermoscopic branch through the continuous cross-modality interaction. To sum up, we finally use $f_{der}$ and $f_{meta}$ for TFormer to get a better performance of lesion classification.

\subsection{Comparisons with State-of-the-art Methods}

\begin{table*}
\renewcommand{\arraystretch}{1.3}
\caption{\label{SOTAs} Comparisons of state-of-the-art methods (accuracy \%). }
\begin{threeparttable}
\resizebox{\textwidth}{!}{
\begin{tabular}{c|cccccccc|c}
\hline\hline
Method                         & DIAG                                 & PN                                   & BMV                                & VS                          & PIG                                & STR                                  & DaG                                  & RS                                   & {\color[HTML]{CB0000} avg.}          \\ \hline\hline
Inception-unbalanced & 68.4                                 & 68.1                                 & 87.6                               & 81.3                        & 65.6                               & 75.9                                 & 56.7                                 & 78.2                                 & 72.7                                 \\
Inception-balanced   & 70.8                                 & 68.9                                 & 87.3                               & 81.5                        & 64.8                               & 75.7                                 & 60.3                                 & 78.2                                 & 73.4                                 \\
Inception-combine    & 74.2                                 & 70.9                                 & 87.1                               & 79.7                        & 66.1                               & 74.2                                 & 60                                   & 77.2                                 & 73.7                                 \\
MmA                            & 70.6                                 & 65.6                                 & 83                                 & 75.7                        & 61.3                               & 69.4                                 & 59.2                                 & 73.9                                 & 69.8                                 \\
EmbeddingNet                   & 68.6                                 & 65.1                                 & 84.3                               & 82.5                        & 64.3                               & 73.4                                 & 57.5                                 & 78.0                                 & 71.7                                 \\
TripleNet                      & 68.6                                 & 63.3                                 & 87.9                               & 83.0                        & 67.3                               & 74.4                                 & 61.3                                 & 76.0                                 & 72.7                                 \\
HcCNN                          & 69.9                                 & 70.6                                 & 87.1                               & \textbf{84.8}               & 68.6                               & 71.6                                 & 65.6                                 & 80.8                                 & 74.9                                 \\
FusionM4Net-FS         & 75.6$\pm$1.2            & 67.6$\pm$1.8            & 87.9$\pm$0.7          & 82.1$\pm$0.5   & \textbf{71.5$\pm$0.7} & 73.8$\pm$0.6            & 60.2$\pm$0.5            & 81.9$\pm$1.0            & 75.1$\pm$0.4            \\
FusionM4Net-SS         & 77.6$\pm$1.5            & 69.2$\pm$1.5            & \textbf{88.5$\pm$0.4} & 81.6$\pm$0.5   & 71.3$\pm$1.3          & 76.1$\pm$1.1            & 64.4$\pm$1.3            & 81.4$\pm$0.8            & 76.3$\pm$0.7            \\
TFormer(${[}1,1,1,1{]}$)         & 78.48$\pm$0.35          & 73.08$\pm$0.32          & 88.13$\pm$0.21        & 81.77$\pm$1.09 & 69.37$\pm$1.15        & 76.03$\pm$0.98          & 64.64$\pm$0.52          & 81.44$\pm$0.83          & 76.62$\pm$0.11          \\
TFormer(${[}1,1,2,1{]}$\_WS)         & \textbf{79.49$\pm$0.54} & \textbf{74.34$\pm$0.93} & 86.67$\pm$0.95        & 83.03$\pm$0.55 & 70.29$\pm$1.67        & \textbf{76.71$\pm$1.64} & \textbf{66.85$\pm$0.75} & \textbf{82.11$\pm$1.45} & \textbf{77.40$\pm$0.59} \\ \hline\hline
\end{tabular}}
\begin{tablenotes}
\footnotesize
\item $[i,i,i,i]$ is used to indicate the settings of HMT blocks in each stage of TFormer, while WS means the feature extractors using \\the weight sharing strategy.
\end{tablenotes}
\end{threeparttable}
\end{table*}

To show the superiority of our method, we compare our TFormer with other state-of-the-art methods for skin disease multi-modality classification shown in Table \ref{SOTAs}, including Inception-based methods (Inception-unbalanced, Inception-balanced and Inception-combine) \cite{kawahara2018seven}, MmA \cite{ngiam2011multimodal}, TripleNet \cite{ge2017skin}, EmbeddingNet \cite{yap2018multimodal}, HcCNN \cite{bi2020multi} and FusionM4Net  \\(FusionM4Net-FS and FusionM4Net-SS) \cite{tang2022fusionm4net}. Except for our method, the rest results are all obtained from the papers of \cite{kawahara2018seven,bi2020multi,tang2022fusionm4net}. Besides, we use $[i,i,i,i]$ to represent the settings of HMT blocks in each stage of TFormer (for example, $[1,1,1,1]$ means the number of HMT in each stage is 1) and WS means the image feature extractors using the weight sharing strategy. From the results, our method outperforms the contrastive methods on avg. and accuracy on more labels. These algorithms can be divided into two categories: Two-stage method FusionM4Net-SS and other end-to-end methods. The results show that the proposed TFormer has significant advantages among the end-to-end methods. Avg. of TFormer ($[1,1,1,1]$) increases by 1.51\% than that of the better method FusionM4Net-FS, while TFormer ($[1,1,2,1]$\_WS) significantly increases by 2.3\% . Meanwhile, for two-stage method FusionM4Net-SS, our TFormer ($[1,1,1,1]$) achieves a comparable avg.. However, FusionM4Net-SS must not only train CNNs and SVM cluster but also search the decision weights of each classifier on the validation set. Compared with other end-to-end methods, the steps are cumbersome and require more computing resources. In comparison, TFormer $([1,1,2,1])$\_WS is a one-stage method that improves avg. by 1.1\%.

\begin{table*}
\caption{\label{Details} Comparisons of state-of-the-art methods via different metrics.}
\renewcommand{\arraystretch}{1}
\begin{threeparttable}
\resizebox{\textwidth}{!}{
\begin{tabular}{c|c|cccccccc|c}
\hline\hline
                                       &                       &                                       &                                       &                                       &                                       &                                       &                                       &                                       &                                       &                                       \\
\multirow{-2}{*}{Method}               & \multirow{-2}{*}{Metric} & \multirow{-2}{*}{DIAG}                & \multirow{-2}{*}{PN}                  & \multirow{-2}{*}{BMV}                 & \multirow{-2}{*}{VS}                  & \multirow{-2}{*}{PIG}                 & \multirow{-2}{*}{STR}                 & \multirow{-2}{*}{DaG}                 & \multirow{-2}{*}{RS}                  & \multirow{-2}{*}{AVE}                 \\ \hline\hline
                                       & SEN                   & 40.80                                 & 63.90                                 & 72.95                                 & 40.60                                 & 48.83                                 & 56.17                                 & 53.70                                 & 63.30                                 & 55.03                                 \\
                                       & SPE                   & 87.74                                 & 83.33                                 & 72.95                                 & 73.03                                 & 77.67                                 & 80.13                                 & 77.27                                 & 63.30                                 & 76.93                                 \\
                                       & PRE                   & 55.24                                 & 66.97                                 & {\color[HTML]{32CB00} \textbf{83.05}} & 45.70                                 & 62.20                                 & {\color[HTML]{32CB00} \textbf{72.67}} & 56.67                                 & 75.40                                 & 64.74                                 \\
\multirow{-4}{*}{Inception-unbalanced} & F1-score              & 43.92                                 & 63.88                                 & 76.39                                 & 61.23                                 & 48.04                                 & 60.15                                 & 53.77                                 & 64.96                                 & 59.04                                 \\ \hline
                                       & SEN                   & 46.00                                 & 65.37                                 & 68.90                                 & 42.53                                 & 52.63                                 & 61.30                                 & 58.40                                 & 73.20                                 & 58.54                                 \\
                                       & SPE                   & 89.34                                 & 83.87                                 & 78.90                                 & 75.60                                 & 78.73                                 & 82.97                                 & 79.43                                 & 73.20                                 & 80.26                                 \\
                                       & PRE                   & 65.08                                 & 67.43                                 & 79.50                                 & {\color[HTML]{32CB00} \textbf{65.33}} & 61.27                                 & 68.73                                 & 59.63                                 & 58.85                                 & 65.73                                 \\
\multirow{-4}{*}{Inception-balanced}   & F1-score              & 50.25                                 & 65.64                                 & 73.62                                 & 47.10                                 & 53.00                                 & 63.98                                 & 58.37                                 & 64.88                                 & 59.60                                 \\ \hline
                                       & SEN                   & 60.42                                 & 68.03                                 & {\color[HTML]{FE0000} \textbf{83.35}} & 49.30                                 & 55.50                                 & 63.87                                 & 58.97                                 & 73.65                                 & 64.14                                 \\
                                       & SPE                   & 91.06                                 & 85.07                                 & {\color[HTML]{3531FF} \textbf{83.35}} & 78.33                                 & 79.80                                 & 79.60                                 & 79.77                                 & 73.65                                 & 81.33                                 \\
                                       & PRE                   & 69.66                                 & 69.33                                 & 78.70                                 & 54.37                                 & 57.73                                 & 62.83                                 & 59.47                                 & 71.60                                 & 65.46                                 \\
\multirow{-4}{*}{Inception-combine}    & F1-score              & 63.40                                 & 68.31                                 & \textbf{80.63}                        & 50.62                                 & 56.26                                 & 63.33                                 & 58.60                                 & 72.40                                 & 64.19                                 \\ \hline
                                       & SEN                   & 57.44                                 & 68.20                                 & 79.15                                 & 50.03                                 & 61.83                                 & 62.70                                 & 58.77                                 & 71.70                                 & 63.73                                 \\
                                       & SPE                   & 92.06                                 & 85.10                                 & 79.15                                 & 80.07                                 & 82.33                                 & 82.70                                 & 79.77                                 & 71.70                                 & 81.61                                 \\
                                       & PRE                   & 63.30                                 & 69.07                                 & 78.50                                 & 58.53                                 & 69.43                                 & 65.30                                 & 59.77                                 & {\color[HTML]{32CB00} \textbf{82.60}} & 68.31                                 \\
\multirow{-4}{*}{FusionM4Net-FS}       & F1-score              & 57.54                                 & 68.39                                 & 78.82                                 & 52.61                                 & 64.33                                 & 63.82                                 & 58.65                                 & 74.57                                 & 64.84                                 \\ \hline
                                       & SEN                   & 54.62                                 & 68.43                                 & 79.20                                 & 42.40                                 & 54.43                                 & 64.80                                 & {\color[HTML]{FE0000} \textbf{64.20}} & 69.75                                 & 62.23                                 \\
                                       & SPE                   & 92.26                                 & 85.17                                 & 79.20                                 & 75.07                                 & 79.50                                 & 84.53                                 & 82.37                                 & 69.75                                 & 80.98                                 \\
                                       & PRE                   & {\color[HTML]{32CB00} \textbf{70.64}} & 69.53                                 & 82.25                                 & 62.57                                 & {\color[HTML]{32CB00} \textbf{69.97}} & 71.33                                 & {\color[HTML]{32CB00} \textbf{65.20}} & 79.65                                 & {\color[HTML]{32CB00} \textbf{71.39}} \\
\multirow{-4}{*}{FusionM4Net-SS}       & F1-score              & 57.70                                 & 68.67                                 & 80.58                                 & 42.88                                 & 56.64                                 & 67.35                                 & \textbf{63.34}                        & 72.26                                 & 63.68                                 \\ \hline
                                       & SEN                   & {\color[HTML]{FE0000} \textbf{73.46}} & {\color[HTML]{FE0000} \textbf{74.46}} & {\color[HTML]{000000} 79.77}          & {\color[HTML]{FE0000} \textbf{65.76}} & {\color[HTML]{FE0000} \textbf{67.51}} & {\color[HTML]{FE0000} \textbf{72.18}} & 64.02                                 & {\color[HTML]{FE0000} \textbf{80.98}} & {\color[HTML]{FE0000} \textbf{72.27}} \\
                                       & SPE                   & {\color[HTML]{3531FF} \textbf{93.71}} & {\color[HTML]{3531FF} \textbf{87.55}} & 79.77                                 & {\color[HTML]{3531FF} \textbf{86.27}} & {\color[HTML]{3531FF} \textbf{82.70}} & {\color[HTML]{3531FF} \textbf{87.06}} & {\color[HTML]{3531FF} \textbf{82.64}} & {\color[HTML]{3531FF} \textbf{80.98}} & {\color[HTML]{3531FF} \textbf{85.09}} \\
                                       & PRE                   & 64.94                                 & {\color[HTML]{32CB00} \textbf{74.65}} & 77.90                                 & 60.50                                 & 66.75                                 & 69.53                                 & 62.70                                 & 75.49                                 & 69.06                                 \\
\multirow{-4}{*}{TFormer($[1,1,2,1]$\_WS)}                 & F1-score              & 66.57                                 & \textbf{74.51}                        & 78.77                                 & \textbf{62.77}                        & \textbf{66.97}                        & \textbf{70.73}                        & 62.93                                 & \textbf{77.50}                        & \textbf{70.09}                        \\ \hline\hline
\end{tabular}}
\begin{tablenotes}
\footnotesize
\item All values take the best results among different methods. The highest value of SEN, SPE, PRE and F1-score are highlighted in {\color[HTML]{FE0000}red}, {\color[HTML]{3531FF}blue}, {\color[HTML]{32CB00}green} and black.
\end{tablenotes}
\end{threeparttable}
\end{table*}

To further display the superiority over other methods. We use additional evaluation metrics including SEN, SPE, PRE and F1-score for comparisons. The data given in the Table \ref{Details} are the average values of each kind of labels, which are calculated by the values given in the papers \cite{kawahara2018seven,tang2022fusionm4net}. F1-score is the arithmetic mean of SEN and PRE, which is used to evaluate the algorithms when the changes of SEN and PRE in different directions and the term AVE in Table refers to the average value of all labels. According to experimental results, our method achieves superiority on most labels in terms of the SEN, SPE, and F1-score metrics. Notably, the SEN value improving by 8.13\% compared to the second place (Inception-combine). Although our PRE value comes the second place, it is only 1.3\% lower than the first place (FusionM4Net-SS). Taking both SEN and PRE into consideration, F1-sore of TFormer get the highest 70.09\%. Furthermore, we also draw radar charts of all multi-label categories (24 categories in total) shown in Figure \ref{radar_chart}, where the paper of HcCNN method only gives the value of metrics in specific labels. So the ungiven information is replaced by the origin in the center of the subgraph. The purpose of a radar chart is to show the comprehensive performance of the algorithm under multiple evaluation indexes, since we use 4 metrics in the comparison, our chart is quadrilateral. A better algorithm should take over a larger area in the chart, thus we can see from Figure \ref{radar_chart} in appendix that our method has a more balanced distribution of multiple indicators in most of the categories.

\section{Conclusions}
In this paper, we propose TFormer, a throughout fusion architecture for MSLD. Considering the issue of insufficient information fusion existing among the previous methods, we leverage the transformer structure instead of deep convolutions which can bring in more representative features in the shallow layers. We then design a stack of HMT blocks which can fully aggregate the information across the image modalities in a stage-by-stage way. Furthermore, we take the gap between modalities of heterogeneous data types into account and adopt a ``divide and conquer" strategy to fuse features of image and meta-data modalities. With the designed MTP block, we can effectively utilize the information of image and non-image modalities. The experimental results on the public available Derm7pt multi-label dataset show that our method surpasses other state-of-the-art approaches.   

\section*{Funding}
This work was supported by the National Natural Science Foundation of China (Nos. 61871011, 62071011 and 61971443).

\section*{Appendix}
\appendix
\setcounter{figure}{0}
\renewcommand{\thefigure}{A\arabic{figure}}
\begin{figure*}
	\centering
	\includegraphics[scale=0.18]{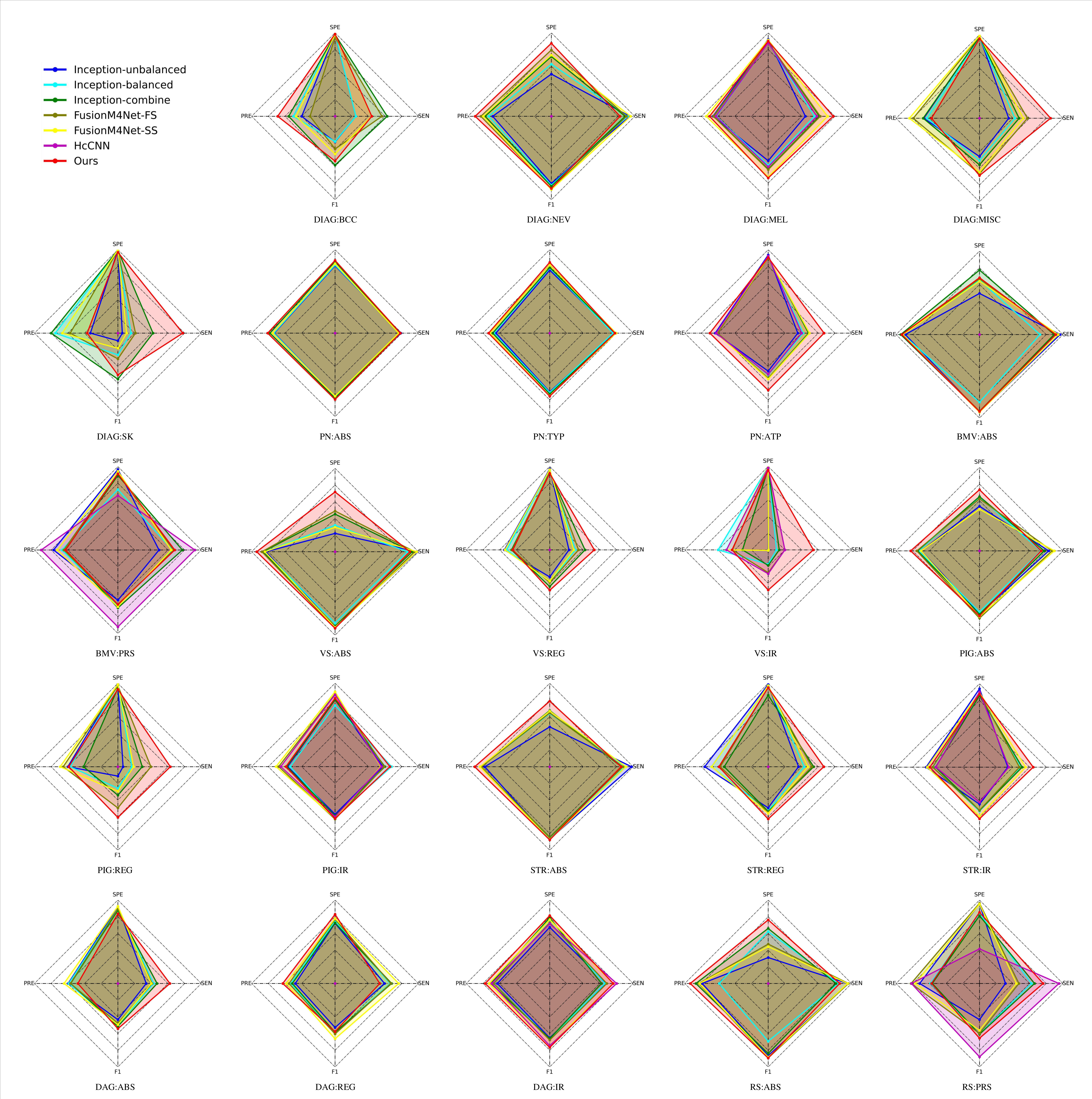}
	\caption{The radar charts of all multi-label categories with the metrics including SEN, SPE, PRE and F1-score.}
	\label{radar_chart}
\end{figure*}
See Figure \ref{radar_chart}.

\bibliographystyle{cas-model2-names}

\bibliography{manuscript}

\end{document}